\documentclass{article}
\usepackage{authblk}
\usepackage{graphicx} 
\usepackage{verbatim}
\usepackage{framed}
\usepackage{hyperref}
\usepackage{biblatex} 
\usepackage{tikz}
\usetikzlibrary{shapes.geometric, arrows}

\title{Towards a Holodeck-style Simulation Game}

\author[1]{Ahad Shams}
\author[1]{Douglas Summers-Stay}
\author[1]{Arpan Tripathi}
\author[1]{Vsevolod Metelsky}
\author[1]{Alexandros Titonis} 
\author[1]{Karan Malhotra}

\affil[1]{Webaverse}

\begin{document}

\maketitle

\usetikzlibrary {graphs}
\begin{figure}
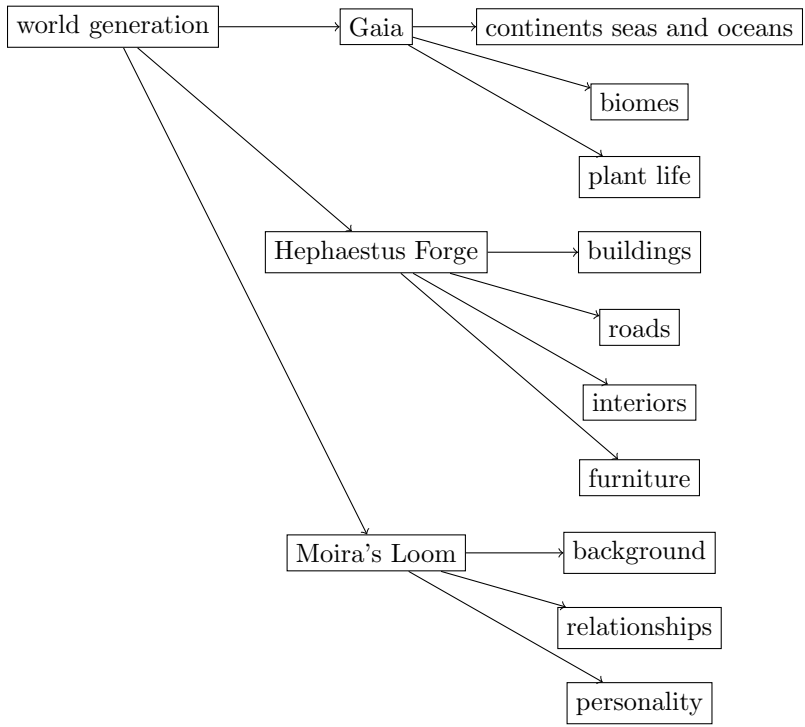

\centering
\tikz [every node/.style = draw]
  \graph [grow right=3.5 cm,
          branch down=1cm] {
    world generation -> {
        Gaia -> {
            continents seas and oceans,
            biomes,
            plant life
        },
        Hephaestus Forge -> {
            buildings,
            roads,
            interiors,
            furniture
        },
        Moira's Loom -> {
            background,
            relationships,
            personality
        }
    }  
};    
\caption{Generating a world is broken down into several modules that work together based on an overall text prompt describing the world to be generated. A variety of procedural generation techniques and AI sprite generation are used to create an entire explorable world, complete with inhabitants who interact with each other and with the player in rich, complex ways.}
\end{figure}

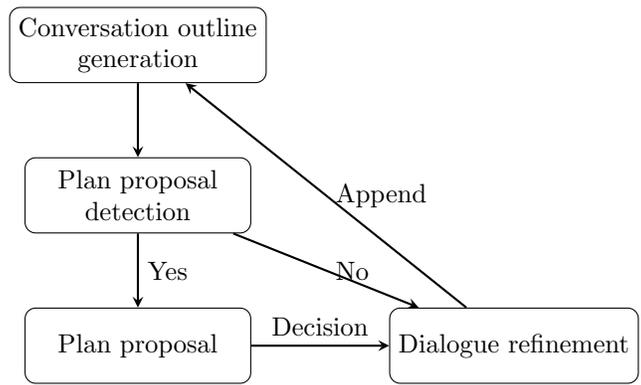
\begin{figure}
\centering
\begin{tikzpicture}[
  startstop/.style={rectangle, rounded corners, minimum width=3cm, minimum height=1cm,text centered, draw=black, align=center},
  arrow/.style={thick,->,>=stealth}
]

\node (start) [startstop] {Conversation outline\\ generation};
\node (in1) [startstop, below of=start, yshift=-1cm] {Plan proposal\\ detection};
\node (out1) [startstop, below of=in1, yshift=-1cm] {Plan proposal};
\node (stop) [startstop, below of=in1, yshift=-1cm, xshift = +5cm] {Dialogue refinement};

\draw [arrow] (start) -- node[midway,right] {}(in1);
\draw [arrow] (in1) -- node[midway,right] {Yes} (out1);
\draw [arrow] (in1) -- node[midway,right] {No} (stop);
\draw [arrow] (stop) -- node[midway,right] {Append} (start);
\draw [arrow] (out1.east) -- node[midway,above] {Decision} (stop.west);

\end{tikzpicture}
\caption{High level architecture of conversation system}
\end{figure}

\section{Abstract}
We introduce Infinitia, a simulation game system that uses generative image and language models at play time to reshape all aspects of the setting and NPCs based on a short description from the player, in a way similar to how settings are created on the fictional Holodeck.  

Building off the ideas of the Generative Agents paper, our system introduces gameplay elements, such as infinite generated fantasy worlds, controllability of NPC behavior, humorous dialogue, cost \& time efficiency, collaboration between players and elements of non-determinism among in-game events.

Infinitia is implemented in the Unity engine with a server-client architecture, facilitating the addition of exciting features by community developers in the future. Furthermore, it uses a multiplayer framework to allow humans to be present and interact in the simulation. The simulation will be available in open-alpha shortly at \href{https://infinitia.ai/}{https://infinitia.ai/} and we are looking forward to building upon it with the community.

\section{Introduction}

in 1987, the TV show \textit{Star Trek: The Next Generation} introduced a novel concept into science fiction called ``The Holodeck." The Holodeck was a multi-player entertainment room that used Artificial Intelligence (AI) to generate a believable proxy of an environment based on verbal instructions. The capabilities of the Holodeck restricted solely by the user's imagination, including intricate details related to inanimate objects and AI-powered characters, that the user could interact with either directly, as if they were other people, or indirectly, by asking the computer to revise the simulation.

The list of technologies needed to realize the Holodeck seems daunting, as anything that exists in the generated world would need to be simulated accurately, as the user does not have any restrictions. However, we are advancing towards making it a possibility with the incorporation of generative AI tools. The current state-of-the-art provides excellent generality over images and text, potentially bringing such a vision within reach. The technology shows a clear and rapid positive trend, courtesy of the democratisation of generative AI technologies. However, generating images is only one small part of generating simulated worlds. Understanding how those images should be arranged in the world, keeping them consistent with one another, simulating their interactions, and bringing the NPCs to life provide enormous challenges.

In this paper, we describe our simulation game, named Infinitia, that uses a combination of Generative AI and procedural generation to create new worlds for every player. It is a top-down-oblique exploration and simulation game. Infinitia generates the terrain, vegetation, buildings, and inhabitants of the world based on the player’s preferences. The game aims to provide a unique and immersive experience for every player, as they can explore and interact with a constantly evolving world that adapts to their choices. Initialising with a one phrase description of the world they want to explore creates every aspect of the game custom tailored for the player.

Beyond simply creating an explorable environment, the system populates the world with AI-powered NPCs, characters with entertaining backstories, interpersonal relationships, strong desires and independent lives that are influenced by every action the player takes in a rich, context-sensitive way. These NPCs have the capabilities to hold conversations, alter their environment to better suit their goals, and adapt their routines based on environmental stimulus.

\section{Background and Related Works}
Procedurally generated game elements have been used since early computer games era, serving as a means to introduce content with reduced developer effort to store all the possibilities in the game. \textit{Beneath Apple Manor} (by Don Worth in 1978) and \textit{Rogue} (by Toy and Wichman in 1980) gave rise to ``roguelike" games that randomly generated a dungeon setting of rectangular rooms connected by passages. Beginning with \textit{Elite} (by Brabensoft in 1984), galaxy exploration games used procedural generation to create many worlds for players to explore; a modern example is \textit{No Man's Sky} (by Hello Games in 2016.) The ``demoscene'' was an early pioneer of many procedural generation techniques. Perlin noise was invented by Ken Perlin in 1983 while working on the movie \textit{Tron}. 

The use of generative neural models for creating games is much more recent. One of the first games to use content generated by neural networks was \textit{Creatures} (by Toby Simpson in 1996). It used the neural network to influence game play, by affecting the way the `creatures' in the game learned to respond to various stimuli. Fantasy Raiders tried using GANs and LSTMs to generate levels in 2017, with mixed results. \textit{AI Dungeon} (Nick Walton in 2019) was the first game to use neural text generation for story and dialogue elements. Generating concept art, assets, and dialogue for games using modern image generation networks (such as Stable Diffusion) and generative language models (such as GPT) is widespread today, although limited by social constraints. Scenario.gg and Ubisoft Ghostwriter for example, provide models for such purposes.

Procedurally generated NPCs have been used in games such as \textit{Watch Dogs: Legion}. In this game, the world contains hundreds of NPCs who have a very shallow description and behaviors. However, when the player singles them out for attention, that backstory is filled in and linked with other NPCs the player has interacted with, giving the impression of a fully simulated social world.

An attempt to simulate realistic human behavior in NPCs by using language models was made in the recent Generative Agents\cite{park2023generative} work, which proposed a vector-based memory retrieval system along with a reflection and planning system for simulating NPCs, however, the setting was limited to a single level environment, one-on-one NPC conversations and a world with no lore of its own. We aim to extend the framework to vast open AI generated worlds, with support for group conversations, distance based perception, time based memory and more control over NPC behavior.    

\section{Game Design}
\textit{Infinitia} focuses on exploration, creation, and interaction. The simulation is primarily a social simulation at this stage of development, modeling the internal thought process and actions of large groups of AI-powered NPCs. 

The underlying architecture supporting the AI-powered NPCs operate independently from the algorithmic components of both the user interface and the game's core functionality, both of which have been fully implemented within the Unity Engine framework, with the exception for the overlapping components such as perception and routines, and thus, can be generalised into settings other than that of our implementation. For instance, the same underlying technology could be used to build isometric, 3D or vector art styles.

\subsection{World Building}

\subsubsection{Creating a Natural Environment with Gaia}
We designed Gaia to be able to create a huge variety of possible game worlds-- contemporary, historical, fantasy or science fiction. Here are just a few example worlds that can be created:
\begin{itemize}
    \item a post-apocalyptic volcanic wasteland
    \item a district of cozy villages in southern France
    \item a harsh world inspired by Norse mythology
    \item a cyberpunk megalopolis, where the trees are all replaced by solar collectors
    \item a haunted country with decaying castles and twisted forests
    \item a world inspired by the sea floor, with houses built of shells and coral trees
    \item a world where everything is made of candy
\end{itemize}
Since our objective is to create a system capable of generating a variety of game worlds without restricting the possibilities, we need a general creation scheme that is customise-able and gives rise to wildly different game appearances. We use a mixture of more traditional procedural generation techniques with novel techniques that use make use of language and image generation.

For many aspects of the world, we create a generic world, and then ask a language model to form analogies between that world and the world the player has described. The analogical asset descriptions generated as text are passed to a text-to-image generator, to create assets that serve the same in-game function as the generic ones, but are appropriate to the described environment. For example, in an ``world made of glass" a ``spreading apple tree" might map to a ``wide glass tree with glass ornaments." But if the apple tree produces apples that are edible, the ornaments produced by the glass tree would also be edible.

To produce a varying game-play experience which rewards exploration, we use traditional procedural generation techniques to produce terrain and assign it varying natural regions. In the generic world, these regions are biomes such as desert and rain forest, but they are mapped to analogical regions appropriate to the new setting.

Perlin noise is a prominent technique for generating terrain in game environments due to its ability to produce visually appealing and organic-looking landscapes with diverse features. The algorithm utilizes coherent gradients and interpolation methods to create terrains characterized by smooth transitions, realistic slopes, and varied topographic elements, resembling natural landscapes. By incorporating fractal properties, Perlin noise successfully simulates terrain patterns observed in nature, including mountains, valleys, cliffs, and plateaus, enhancing the immersion and engagement for players. It strikes a favorable balance between visual quality and computational efficiency, making it well-suited for real-time applications like game development. Its inherent parallelizability facilitates efficient implementation on modern graphics hardware, enabling the generation of terrains at interactive frame rates. Additionally, the computational cost associated with Perlin noise remains relatively low compared to more computationally demanding algorithms, making it an appealing choice for resource-constrained platforms.

We use Perlin noise to generate terrains and to assign precipitation and modify temperature levels. Based on these properties, the ground tiles are assigned and individual plants and other natural objects appropriate to those regions are placed. The sprites for these assets are generated as described in the next section.

\subsubsection{Text-to-image Asset Generation with Daedalus}

While text-to-image generative models have made enormous progress in the last few years, there are still several issues involved in generating new assets to be used in-game without any human supervision in the process.

\begin{figure}
    \centering
    \includegraphics[width=\linewidth]{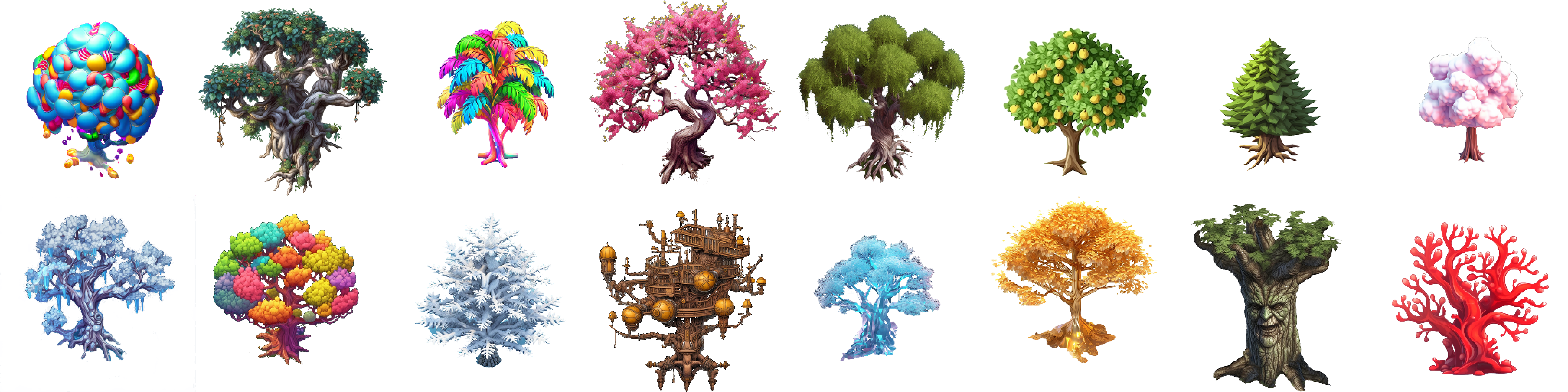}
    \caption{Generating natural world game assets with a custom Stable Diffusion model allows us to adapt the appearance of the setting to a wide variety of descriptions.}
    \label{fig:trees}
\end{figure}

\textbf{Consistency of Style}

A ``pixel-art" game means, roughly, a game in the style of the 1980s and 1990s when resolution was low and views of objects were created by hand by artists deliberately choosing the color of each pixel. Using Stable Diffusion to generate ``pixel-art" results in a variety of pixel scales and artist styles, from heroic fantasy to cute. We needed to both train new models and use prompt engineering to get a unified style of outputs and then apply post-processing to create a unified palette and pixel scale for each sprite.

\textbf{Consistency of Scale}

Stable Diffusion is generally capable of generating images in a narrow range of sizes around 512x512. However, we needed the ability to generate objects as small as mosquitoes and as large as a factory. This means generating at one scale, automatically resizing, and then pixelating as appropriate for the new scale. We found that LLMs were able to provide a reasonable estimate of size to enable automatic resizing.

\textbf{Consistency of Viewpoint}

The viewpoint camera in our game is placed so that the top and front of objects are visible, but not the sides. This is known among game designers as a ``top-down oblique" viewpoint. Attempting to generate images from this viewpoint often resulted in either true top-down images, front-view images, or isometric images. We tried several fixes for this, including careful prompting and (as they became available) ControlNet techniques to constrain generation. Finally, we trained our own models on hundreds of examples. For small objects, the in-game style is often iconic, meaning the viewpoint can vary between front, top, and isometric and it is simply accepted as a game convention for legibility. For buildings, we found that including the phrase ``large roof" helped nudge the model towards showing building roofs as well as their facades.

\textbf{Creativity of Generation and Consistency with Prompt}

Two methods for enforcing style and viewpoint are fine-tuning and modifying the prompt and negative prompt. However, both of these come with a trade-off with the creativity of the image generation model. We want the image generated to be original for each new world, but the more restricted the generation is, the more similar they look. Generative models often ignore some part of a long prompt. If the part ignored is the part that varies with different worlds, then the generated asset will not be unique to that world. In order to prevent this, the fine-tuning examples must cover a broad range of possibilities and the prompt should be kept as simple as possible while preserving the necessary properties.

Throughout this paper, whenever we refer to generating sprites with a prompt, two methods are used in the game. The first is direct generation from a prompt, as described above. However, the generation process is not always successful, failing in style, viewpoint, background removal, or other ways. One way of dealing with this is to present the user with multiple generations and let them choose the one that works best (as is done in, for example, the image generation platform Midjourney). The other approach is to have a large library of pre-generated assets, and use the prompt to retrieve the asset that is most semantically similar to the prompt. This allows us to guarantee a level of quality by visually inspecting each asset before adding it to the library. We use CLIP to match the text of the prompt to the images in the library.  

\textbf{Background Removal}

The current text-to-image generators only generate RGB images without a transparency channel. For a generated image to be used as a game asset, the background and foreground needs to be separated. There are broadly two approaches for background removal: neural approaches and traditional image processing.

Neural background subtraction can sometimes have amazing results. But most current methods expect the background to look like the background of a photograph: sky, trees, buildings, roads, grass. This makes them fail on the simple gradient backgrounds in generated images, which are mostly out-of-distribution for the background removal networks. We use a neural background remover trained on the kind of subtle gradients with or without shadows we see in our generated images. 

\subsubsection{Creating the Built Environment with Hephaestus's Forge}

Without knowing beforehand what type of world will be generated, realistic room, building, and town layouts would be difficult to generate: cultural differences lead to different layouts in a way that is not easy to characterize, and neither language models (which have impaired spatial sense) nor image models (which, even if they have seen floor plans and city plans from multiple cultures, cannot generate labeled images) are good at this. Fortunately, video game depictions of layouts are quite simplified from reality and are more consistent than reality. In typical top-down games, most buildings only have a few rooms, and settlements usually have between five buildings (for a village) and thirty buildings (for a large city). Additional considerations are that the facade of a top-down building must be the side facing the player, and the facade entrance cannot be hidden behind another building. These constraints mean that all buildings will either tend to be placed on horizontal streets or isolated from each other.

\begin{figure}
    \centering
    \includegraphics[width=\linewidth]{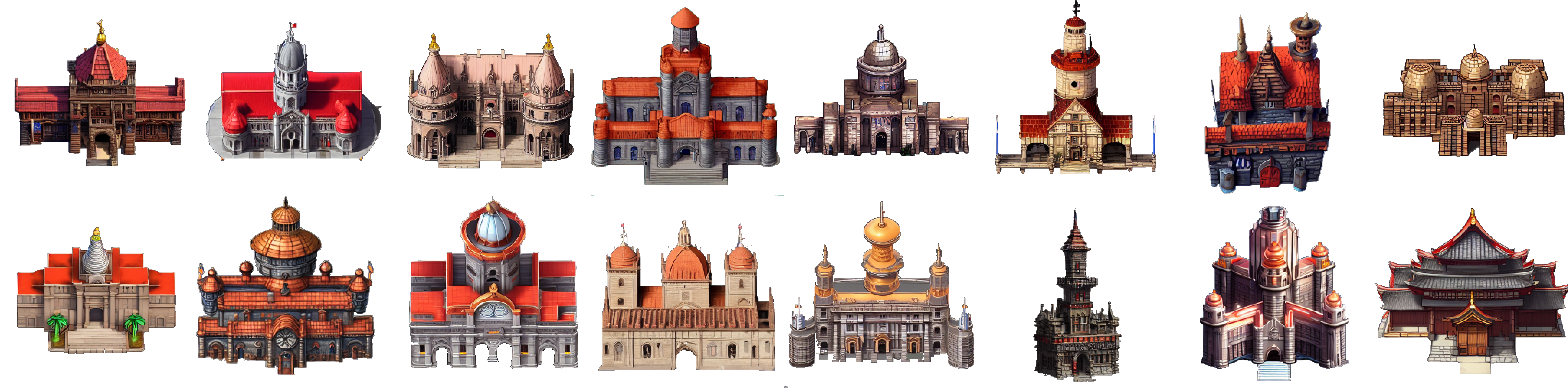}
    \caption{A building fulfilling the function of a ``city hall" in the game, but skinned for a variety of different worlds with radically different appearances. Despite their differences, each building has a consistent point of view, entrance location, artistic style, and function in game. }
    \label{fig:cityhall}
\end{figure}

\begin{figure}
    \centering
    \includegraphics[width=\linewidth]{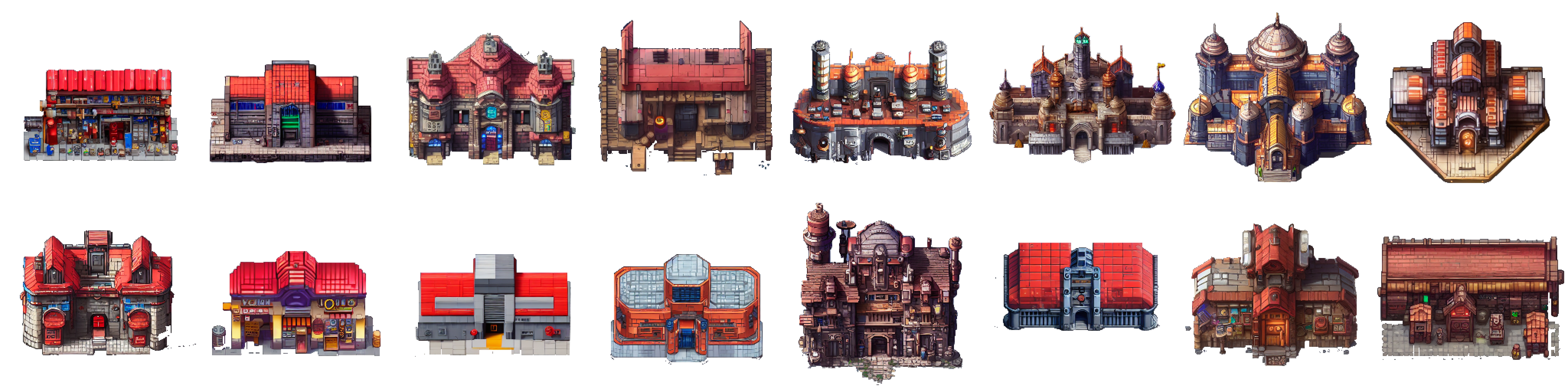}
    \caption{Buildings from the same world (in this case, a future science-fiction world) have a consistent style but are clearly different in function as shops, homes, palaces, prisons, and so forth.}
    \label{fig:future}
\end{figure}

\begin{figure}
    \centering
    \includegraphics[width=\linewidth]{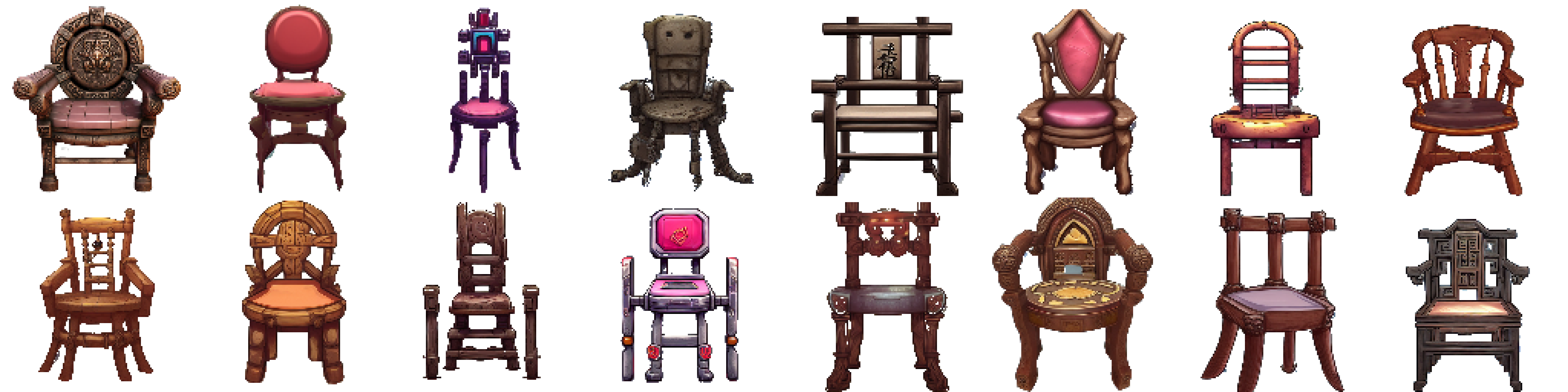}
    \caption{The appearance of each item of furniture within a world is modified to match the setting of the world.}
    \label{fig:chairs}
\end{figure}
\subsubsection{Road Networks}

We have not added vehicles to our worlds, so it makes sense for roads to be irregular dirt walking paths. In order to connect all the buildings of the world by road, we use a fast, approximate traveling-salesman-problem solver to find a route connecting a set of buildings. We then use the A* pathfinding algorithm to find the lowest-cost path through the terrain to follow this route, avoiding impassible areas.

\subsection{Weaving Fates with Moira's Loom}
Once we have generated the world, the next step is to populate it with convincing proxies of life. We took the Generative Agents \cite{park2023generative} work as an inspiration for the implementation of our custom framework, Moira's Loom. Here are a few things each NPC is capable of with this system:
 \begin{itemize}
     \item simulating complex human behavior such as playing a role in a family
     \item conversing with a group while recalling relevant memories on-the-fly
     \item going about their routines and deviating from them based on observed changes in surroundings
     \item forming and tracking plans with other NPCs
     \item forming relationships
     \item reflecting on their day and evolving their personality traits based on insights \item displaying emergent social behavior
 \end{itemize}

We are also working towards adding several other features like long-term goals, free-exploration, crafting, co-ordination, combat and more to make the NPCs truly life-like. This also includes other features like worlds where believable proxies of popular media characters interact with each other.  

\subsubsection{Creating Families and Initialising Relationships}
Based on factors like custom world lore and target world population, Hephaestus's Forge handles the generation and placement of appropriate residences, workplaces and other necessary buildings to accommodate the NPCs. 

Based on the world lore and structure, Moira's Loom generates the background lore of each family, as well as all the associated details like individual lore, character traits, role in the family, seed memories connecting each family member to the other.  

In addition the relationships established within each family, we also connect acquaintances from common workplaces via seed memories and ensure overall connectedness of the social network within a few degrees of separation. The seed-memories utilise the commonalities between the NPCs and facts about the world structure to make them believable.

\subsubsection{Bringing NPCs to Life with Pygmalion}
Based on the NPC's individual and family lore, personality description and layout of the world, continuous and coherent routines are generated. While going about their routines, the NPCs can choose to deviate from the current routine based on the urgency of their observations. The candidate objects for the routines are restricted to the objects accessible to the NPC, i.e, at their residence and workplaces.  

\subsubsection{Perception}
Upon every new observation of the surroundings, the NPC can choose to perform various actions that require them to deviate from their current activity, such as responding to anomalous observations or deciding to talk to another NPC. The NPCs will strike up a conversations based on the actions of other NPCs or react to an emergency. For instance, upon observing an object in a burning state, they may attempt to go for help or try to put it out.

\subsubsection{Conversations}
Each NPC can engage or join ongoing conversations with single or multiple NPCs, which is guided by a polling system that decides the next speaker based on the relevant memories recalled by each NPC with the latest context. The NPCs continue to observe their surroundings and may address the changes in the surroundings in their dialogues. The conversation is summarised as a concise memory for each PC from that PC's point-of-view, subject to their own personality and background.

\subsubsection{Coordination and Planning}
Based on the flow of conversation, NPCs can choose to propose plans. The addressed NPCs for the plan can accept or reject it based on various factors. The plan then gets scheduled as a common routine entry for the involved NPCs. Each NPC also has a memory of accepting/rejecting the plan which they may discuss with other NPCs in another conversation. This system involves maintaining a buffer of plans and accommodating plans on designated days while preserving the overall routine.   

\subsubsection{Reflections}
The NPCs reflect upon their day, including routine events, conversations, and observations. They draw high-level insights based on it, and the system decides the evolution of their personalities based on their experiences, which allows emergent behavior and also affects their future routines. NPCs also have the option to reconsider attending their plans during the reflection.

\subsubsection{Building Relationships}
Other than the existing family members, NPCs also have relationships with their colleagues in form of seed memories that are related to their common workplace. Additionally, NPCs can show emergent social behaviors based on their personality traits, approaching unknown NPCs and forming relationships based on conversations, plans and other latent factors.  

\subsubsection{Commands}

In a mechanic we call Word-of-God, The user can assume the role of the NPC's subconscious and dictate actions in natural language. These are simplified into actionable steps, such as engaging a set of NPCs with an intent for guiding a conversation, performing custom actions/routine entries that other NPCs can witness and react to, or making custom plans at available locations in the world. This framework of dividing the user commands into steps was motivated by the SayCan \cite{ahn2022can} framework, this allows for execution of complex instructions that require thorough planning/various steps to execute believably. 

The user also has the option to interview the NPCs as if they were a character in the simulation, which uses a component of the NPC conversation system, including recalling memories on-the-fly. At the end of the interview, the user can choose whether the NPC remembers the conversation.

Some of the things a user can command an NPC to do are:
\begin{itemize}
    \item start a conversation with another existing NPC
    \item schedule an action for the future at an existing location in the world
    \item propose a plan to other NPCs
\end{itemize}
These actions are observable by other NPCs and has provides potential to lead to interesting interactions. 

\section{Emergent gameplay}
The NPC system, generated world and the creativity of LLMs facilitate emergent behavior with interesting observations that neither we nor the players would anticipate. We will soon release an open-alpha of the simulation at \href{https://infinitia.ai/} for users to experience firsthand and build upon future explorations with the community.
Here are some examples of the way the system responds creatively and appropriately to new situations.

\begin{figure}
    \centering
    \includegraphics[width=0.8\linewidth]{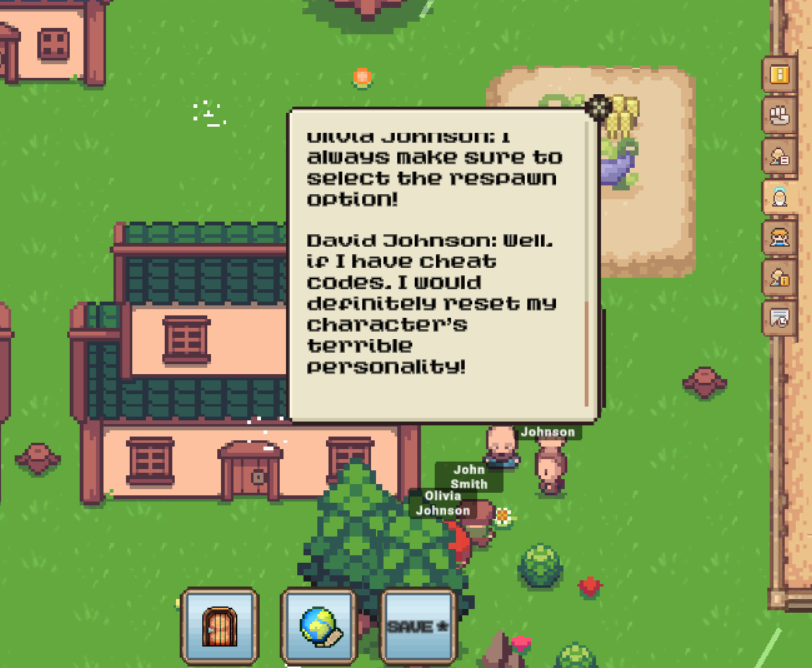}
    \caption{Using Word-of-god, players can instigate interesting conversations between the NPCs on-command}
    \label{fig:welcome}
\end{figure}

\begin{figure}
    \centering
    \includegraphics[width=0.8\linewidth]{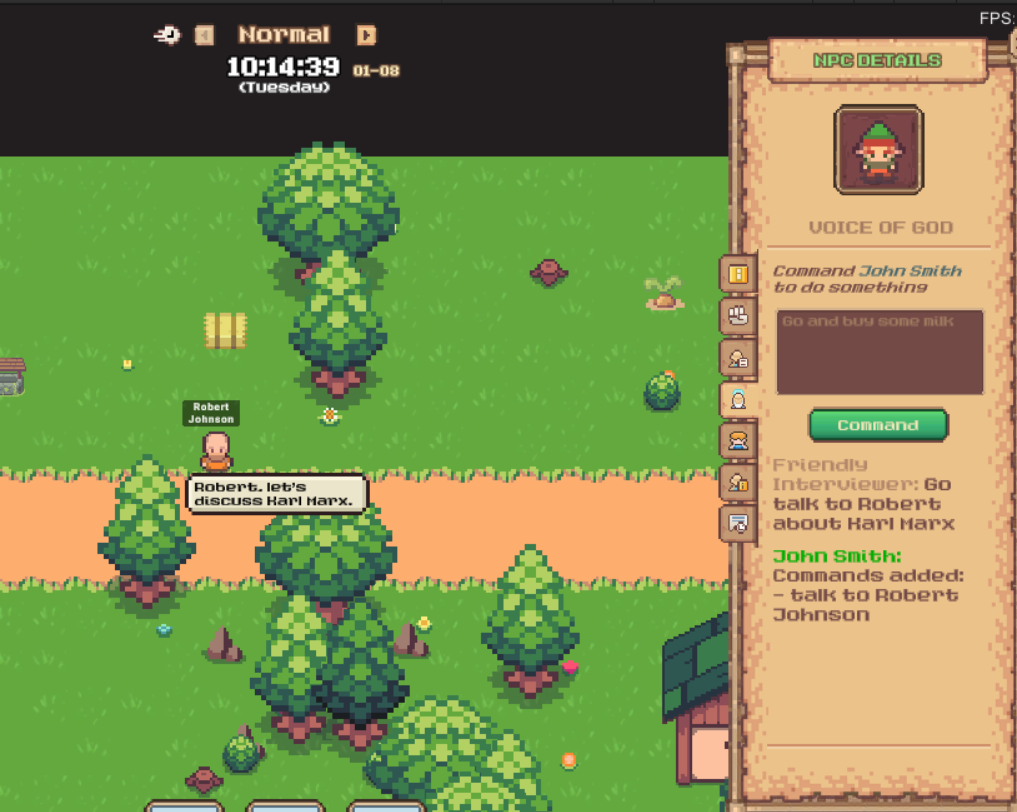}
    \caption{Upon introduction of the factions system in the future, we expect players to jailbreak our underlying LLMs and invent unforeseen outcomes, like communist NPCs}
    \label{fig:welcome}
\end{figure}

\begin{minipage}{\linewidth}
\begin{framed}
Michael Johnson: I once tried to cook a shoe. It tasted better than my cooking.

John Smith: I once tried to cook a pineapple. It tasted like disappointment.

Michael Johnson: Cooking is like a never-ending experiment. Who knows what flavors I'll discover next?

John Smith: I once tried to cook a rubber duck. It tasted like a science experiment gone wrong. Hey Michael, I have a crazy idea! How about we have a cooking competition using unusual ingredients?

Michael Johnson: Sure, John! I'm up for the challenge. Let's have a cooking competition with unusual ingredients!

John Smith: Let's make it really interesting. How about we have a cooking competition using exotic fruits from House2?

Michael Johnson: Sounds like a fun and exciting challenge, John! I'm definitely in for the cooking competition with exotic fruits!

John Smith: How about we include pineapple in the cooking competition? We can use the pineapple from House2 as an alternative to durian.

Michael Johnson: Sure, John! I'm in for the cooking competition with exotic fruits. It's going to be a thrilling experience!
\end{framed}
\end{minipage}

\begin{center}
    Two NPCs decide to have a cooking competition. They discuss using items and locations that exist in the world and make plans around them.
\end{center}

\begin{figure}
    \centering
    \includegraphics[width=0.8\linewidth]{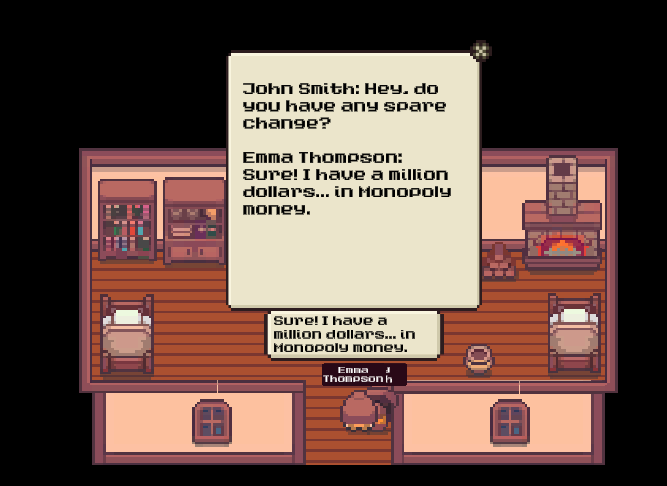}
    \caption{NPCs use context to react appropriately to novel situations. Here, the NPC decides to be quirky.}
    \label{fig:saynothing}
\end{figure}

\begin{figure}
    \centering
    \includegraphics[width=0.9\linewidth]{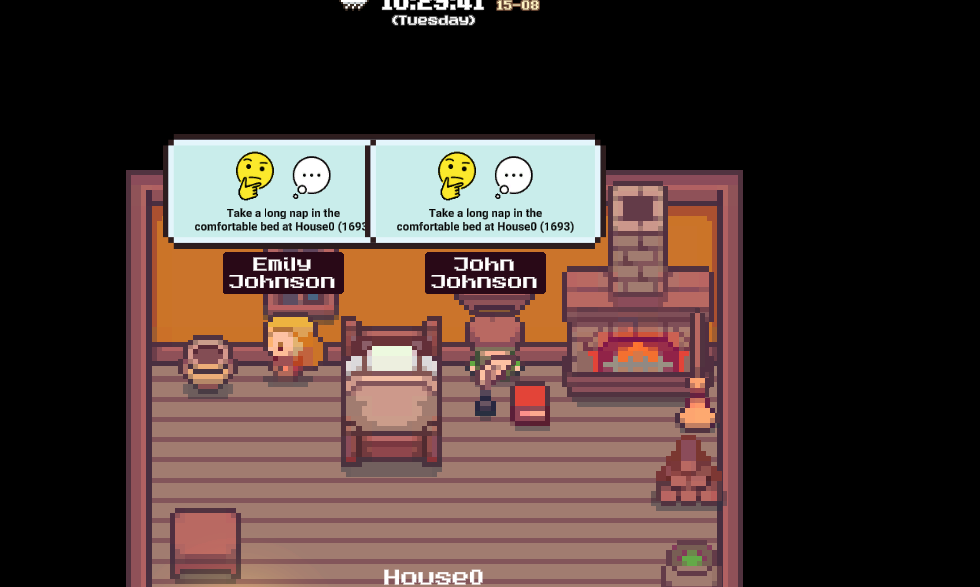}
    \caption{NPCs make plans with each other, recording them in their schedules and altering their behavior accordingly. Here the couple earlier decided to make a couple-y plan and followed it through.}
    \label{fig:plans}
\end{figure}

\begin{figure}
    \centering
    \includegraphics[width=0.9\linewidth]{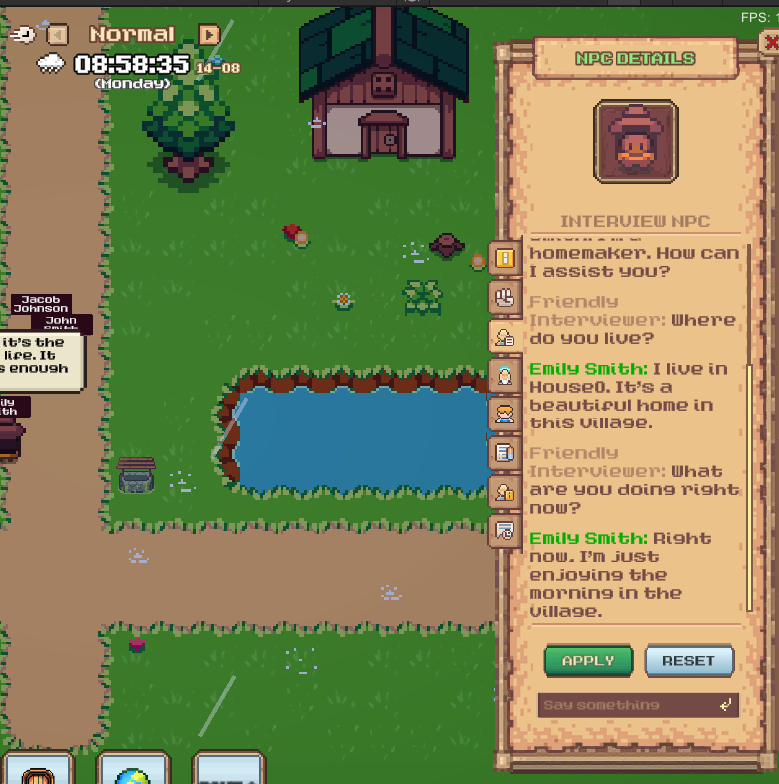}
    \caption{NPCs can also be engaged through the interview system.}
    \label{fig:plans}
\end{figure}

\begin{framed}
Today was a fulfilling day. I love taking care of my family and cooking for them brings me so much joy. It was lovely to have conversations with David during dinner and share stories. Michael, as always, was in the kitchen with me, eager to help. It reminded me of my dream to open a small bakery in the village. I can't wait to see that come true.
\end{framed}
\begin{center}
    At the end of each day, NPCs reflect on the experiences they have had during that day, creating a summary memory that can be recalled later, to give them context to understand conversations and make plans.
\end{center}

\section{Challenges}
\subsection{LLM Hallucinations}
Currently, when the LLM has no relevant context related to the conversation, it improvises to continue it. However, the improvisations are frequently unrelated to the lore of the generated world, which causes the believability of the generated world to suffer. This is related to the broader challenge in the domain of Natural Language Processing(NLP) called LLM hallucinations. Retrieval Augmented Generation (RAG) is a method to mitigate this challenge, however, there are issues with coherance in the generations with this method and we look forward to how the AI research community addresses this challenge in the future. 

\subsection{Context Length}
The performance of the system would benefit from massive improvements with enhanced instruction following and reasoning abilities of the LLMs over longer sequences. This is also an active research area in the NLP domain and several works such as RoPE scaling\cite{kaiokendev} are continuing to mitigate this challenge.

\subsection{Model alignment tradeoff}
Due to the alignment of LLMs, the NPCs are rarely rebellious or rude to each other, which makes them less representative of the actual human nature and affects the believability of the simulation. However, the alignment also makes the game suitable for all audiences and maintains an overall positive experience for users interacting with the NPCs via the Interview system. 
The tradeoff would be optimal with an LLM that is autonomous enough to provide NPCs free will, but aligned enough to prevent the darker parts of the training data\cite{jin2023darkbert} of the LLM from surfacing in the model outputs. 

\subsection{Controllability tradeoff}
Our players would virtually be free to prompt NPCs to do or say anything to each other allowed within the game rules and logic, which might include instructions or information that go against the lore of the generated world or the NPC's character description. The optimal tradeoff between the believability of the world and the controllability is currently unknown and we would rely on user feedback to decide the same.

\section{Future Explorations}

\subsection{Autonomous worlds}

The above generalisable system can create vastly different Worlds. In this context, a World is a container for entities and a sufficiently coherent internal ruleset guiding the behavior. When a system of entities and rules comes to life, it becomes a World.

The system should be able to aid the user in setting up the rules of the world which include physics, society and relational structure, intrinsic desires and motivations of different entities, and so forth.

Once these rules are well defined, the simulation can be automated and run for an indefinite period of time. Most of the rulesets for a World are static (such as physics), while other rules can be dynamic and evolving (such as social law). These would require some form of consensus amongst agents in the simulation.

For now, we use the same ruleset for different worlds, but further explorations will be carried out in the future. We look forward to other teams exploring generalisable systems for different Worlds \cite{autoworld} and scenarios.

\subsection{Quest and Story Generation with Calliope}
The simulation described above supports a game focused on creativity, exploration, and building a growing colony. There is a lot of fun to be had in interacting with characters, seeing what happens when you change aspects of the game, generating different worlds, and so forth. 

However, without a guiding story line and quests that lead toward some larger goal, the game could seem aimless. One of our future milestones include pushing beyond current generative quest systems and make use of the storytelling abilities of AI models to generate quests that are unique to each generated world, are fun to play, and contribute to an overall story arc.
There are two main ways to go about this. One is to include the locations, objects, and people in the world in the context of the generator and prompt it to generate quests using these resources. The alternative is to allow the quest generator to add new locations, people, and objects to the game, and rely on our other generators to create them. Both approaches have tradeoffs, and this is still a challenge given the current state of the generative models. The associated system, named Calliope, is still under development and we aim to present it in the near future.

\subsection{Better Generative AI}
\subsubsection{Image Generation Models}
With the recent release of Stable Diffusion-XL, which displays great performance on pixel-art compared to its predecessors, the generated pixel-art assets would be far better using techniques like Dreambooth\cite{ruiz2022dreambooth} and Low-rank Adaption\cite{lora_stable}.

\subsubsection{Large Language Models}
Although the ChatGPT API, more specifically, the function calling feature, was crucial in the development so far, the frequent degradation in generation qualities was a major roadblock in the development, as our ~30 prompts had to be attended to after each degradation. Therefore, we aim to transition to custom finetunes of the recently released LLaMA 2\cite{touvron2023llama} model specialised for function calling, that involves restricting the LLM generations to adhere to regular expressions. There are emerging frameworks like Guidance\cite{guidance} and Outlines\cite{willard2023efficient} that facilitate the restriction of model vocabulary at inference time to adhere to regular expressions and context-free grammars.   

\section{Conclusion}
Infinitia is a work in progress, but we have the foundation that is already demonstrating the incredible potential of bringing together generative image models, language models, and procedural techniques to create dynamic, responsive, living worlds.

This is our initial attempt at the never-ending game; we are striving to harness a new paradigm of intelligence in future versions. We look forward to community feedback at \href{https://infinitia.ai/}{https://infinitia.ai/}

\printbibliography
\end{document}